\documentclass[conference]{IEEEtran}
\IEEEoverridecommandlockouts
\usepackage{amsmath,amsfonts}
\usepackage{algorithmic}
\usepackage{algorithm}
\usepackage{array}
\usepackage[caption=false,font=normalsize,labelfont=sf,textfont=sf]{subfig}
\usepackage{textcomp}
\usepackage{stfloats}
\usepackage{url}
\usepackage{verbatim}
\usepackage{graphicx}
\usepackage{cite}
\usepackage{multirow}
\usepackage{threeparttable}
\usepackage{makecell}
\usepackage{booktabs}
\usepackage{hyperref}
\usepackage{tcolorbox}
\usepackage[utf8]{inputenc}

\title{RAGalyst: Automated Human-Aligned Agentic Evaluation for Domain-Specific RAG}

\author{
    \large 
    Joshua Gao\textsuperscript{*}, 
    Quoc Huy Pham\textsuperscript{*}, 
    Subin Varghese, 
    Silwal Saurav,
    Vedhus Hoskere \\
    [3pt]
    \small \textsuperscript{*}Equal Contribution \\[6pt]
    \large University of Houston \\  \small
    \href{https://joshuakgao.github.io/RAGalyst}{\texttt{https://joshuakgao.github.io/RAGalyst}}
}

\date{July 2025}

\begin{document}

\maketitle


\begin{abstract}
Retrieval-Augmented Generation (RAG) is a critical technique for grounding Large Language Models (LLMs) in factual evidence, yet evaluating RAG systems in specialized, safety-critical domains remains a significant challenge. Existing evaluation frameworks often rely on heuristic-based metrics that fail to capture domain-specific nuances and other works utilize LLM-as-a-Judge approaches that lack validated alignment with human judgment. This paper introduces RAGalyst, an automated, human-aligned agentic framework designed for the rigorous evaluation of domain-specific RAG systems. RAGalyst features an agentic pipeline that generates high-quality, synthetic question-answering (QA) datasets from source documents, incorporating an agentic filtering step to ensure data fidelity. The framework refines two key LLM-as-a-Judge metrics—Answer Correctness and Answerability—using prompt optimization to achieve a strong correlation with human annotations. Applying this framework to evaluate various RAG components across three distinct domains (military operations, cybersecurity, and bridge engineering), we find that performance is highly context-dependent. No single embedding model, LLM, or hyperparameter configuration proves universally optimal. Additionally, we provide an analysis on the most common low Answer Correctness reasons in RAG. These findings highlight the necessity of a systematic evaluation framework like RAGalyst, which empowers practitioners to uncover domain-specific trade-offs and make informed design choices for building reliable and effective RAG systems. RAGalyst is available on our \href{https://github.com/joshuakgao/RAGalyst}{Github}.
\end{abstract}

\section{Introduction} \label{introduction}

Although modern Large Language Models (LLMs) are great synthesizers of information, they still suffer from hallucinations \cite{kalai2025language, huang2025survey}, which refers to the generation of content that appears plausible but is factually incorrect or unsupported by evidence. Mitigating hallucinations is especially important in safety-critical applications (e.g., military operations, cybersecurity, and bridge engineering) where inaccurate information can lead to serious consequences and undermine trust in artificial intelligence (AI) systems~\cite{varshney2023stitch, li2024dawn}. Retrieval-Augmented Generation (RAG) has been widely adopted to mitigate hallucinations by grounding responses in provided context ~\cite{gao2023retrieval, li2024factual}.

 A key advantage of RAG is its ability to provide models with dynamic, inference-time access to private and domain-relevant documents~\cite{gao2023retrieval, izacard2022atlas}. However, leveraging this ability is highly sensitive to several domain-specific variables. Source documents in specialized fields often include out-of-distribution content—such as dense jargon or unconventional formatting—that lies outside the training corpora of the LLM and embedding model, hindering their ability to  interpret the retrieved text. Similarly, the optimal text chunking strategy and retrieved context lengths may differ in different domains due to document structure and the amount of contextual information required to synthesize information. For example, bridge engineering documents may require an understanding of deterioration trends across extended spans of inspection narratives and historical measurements, favoring longer context windows to capture cross-report dependencies.  On the other hand, cybersecurity documents tend to present short but information-dense logs (e.g., a few lines of packet capture) may be sufficient for accurate interpretation, making smaller chunks preferable.

The challenge of adapting RAG systems to specialized domains is compounded by the difficulty of accurately evaluating their performance. Such evaluation requires domain-specific benchmarks and metrics that produce reliable results.

Given the widespread need for RAG systems and the innumerable domains in which they may be used, manually producing benchmark datasets for tuning of parameters can be very challenging. State-of-the-art RAG evaluation frameworks increasingly employ LLMs to generate synthetic Questions and Answers (QAs) from a knowledge base that are then used to evaluate RAG system performance. The quality of these datasets then become critical in the reliability of reported evaluations. Most approaches require human validation in combination with heuristics to ensure dataset quality. Other fully automated generation pipelines like RAGAS \cite{es2023ragas} lack rigorous quality filtering, resulting in an unreliable QA dataset.

Beyond dataset generation, evaluation metrics are also vital to directly estimate RAG system performance. Early works in RAG evaluation that often rely on traditional heuristic metrics, such as Bilingual Evaluation understudy (BLEU) \cite{papineni2002bleu} and Recall-Oriented Understudy for Gisting Evaluation (ROUGE) \cite{lin2004rouge}, operate by measuring the literal overlap of words or phrases between the generated answer and a reference answer. A challenge lies in their inability to capture semantic meaning; a factually correct response that uses different phrasing would be unfairly penalized, while a nonsensical answer that shares keywords might score deceptively high. To overcome the rigidity of these lexical metrics, researchers have begun leveraging LLM-as-a-Judge to better evaluate the semantic nuances of generated answers~\cite{zhu2025rageval, saad2023ares, es2023ragas, liu2023recall, chen2024benchmarking, yu2023reeval}. However, these LLM-based metrics have not yet been rigorously examined for human-alignment. There is a need to develop human-alignment mechanism to LLM as a judge metrics that can help more reliably automate benchmarking and tuning of domain-specific RAG systems.

This paper introduces RAGalyst, an end-to-end human-aligned agentic framework for domain-specific RAG evaluation. Our framework integrates refined LLM-as-a-Judge metrics and fully automated agentic QA benchmark construction to enable rigorous benchmarking and more reliable deployment of RAG systems across diverse domains. We evaluate the human-alignment procedure on two metrics, the Answerabilty metric - used in synthetic QA generation, and the Answer Correctness metric - used in RAG answer evaluation.

The main contributions of this paper are as follows:
\begin{itemize}
    \item We present an automated agentic framework for evaluating domain-specific RAG systems, integrating document preprocessing, reliable synthetic QA generation, embedding model and LLM-based evaluation modules.
    \item We introduce novel human benchmark-aligned prompt-optimized LLM-as-a-Judge formulations for RAG metrics, namely the Answerability and Answer Correctness metrics, which outperform state-of-the-art formulations as judges.  
    \item We evaluate our framework and demonstrate applicability on documents from on three different specialized domains, namely military operations, cybersecurity, and bridge engineering.
\end{itemize}

\section{Related Work} \label{relatedwork}

\subsection{Retrieval-Augmented Generation (RAG)}

RAG addresses hallucinations by integrating a retrieval module that dynamically retrieves relevant textual chunks from a knowledge base during inference~\cite{gao2023retrieval}. Early RAG systems followed a straightforward retrieve-then-generate pipeline~\cite{lewis2021retrievalaugmentedgenerationknowledgeintensivenlp}. Recent advances include modularization for flexible component composition~\cite{gao2024modular}, adaptive retrieval strategies that dynamically adjust retrieval depth~\cite{jeong2024adaptive}, and graph-based reasoning mechanisms for multi-hop information synthesis~\cite{edge2024graphrag}. The Atlas model demonstrated that few-shot learning achieved strong performance even with relatively small parameter counts when using a dense retriever and joint pre-training strategies~\cite{izacard2022atlas}. LongRAG uses large context chunks and leverages long context LLMs to reduce retrieval noise and improve semantic integrity~\cite{jiang2024longrag}. On the other hand, OP-RAG argues that naive use of long-context models may dilute relevant content, and that order-preserving RAG techniques can offer superior efficiency and answer quality by maintaining the original document structure during chunk selection~\cite{yu2024opr}.


\subsection{Domain-Specific RAG}

While general-purpose RAG systems demonstrate promising capabilities in knowledge-intensive tasks, they often underperform in domains where specialized knowledge or jargon is involved. Domain-specific RAG bridges this gap by tailoring both retrieval and generation processes to the target domain. RAFT~\cite{zhang2024raft} proposes Retrieval-Augmented Fine-Tuning, a training procedure that fine-tunes LLMs to handle bad retrieval to enhance robustness and improve downstream QA performance on specialized datasets such as PubMed and HotpotQA. Similarly, Li et al.~\cite{li2024factual} created a synthetic dataset sourced from Carnegie Mellon University's website, and proposed a domain-specific RAG pipeline. Nguyen et al.~\cite{nguyen2024enhancing} show that fine-tuning both the embedding model and the generator significantly improves performance on complex datasets like FinanceBench.

To benchmark and evaluate RAG capabilities in expert domains, Wang et al.~\cite{wang2024domainrag} introduced \textit{DomainRAG}, a comprehensive benchmark for Chinese university enrollment data. Their work identified six critical abilities for domain-specific RAG systems: conversational handling, structural understanding, denoising, multi-document reasoning, time sensitivity, and faithfulness to external knowledge. They demonstrated that current LLMs struggle significantly in a closed-book setting, validating the necessity for domain-specific RAG systems.

\subsection{RAG Evaluation}

RAG performance depends on a complex interplay of components, including retrieval module, document chunking strategies and model prompting~\cite{yu2024evaluation}. Unlike standalone LLMs, RAG systems are highly sensitive to changes in these components, making generalization between tasks and settings extremely difficult.

Traditional evaluation metrics such as BLEU \cite{papineni2002bleu}, ROUGE \cite{lin2004rouge}, or exact match fail to account for the modular and domain-specific nature of RAG pipelines. Consequently, a variety of frameworks have emerged to address this challenge. RAGEval~\cite{zhu2025rageval} and ARES~\cite{saad2023ares} are prominent reference-free and semi-supervised evaluators that assess context relevance, answer faithfulness, and informativeness without requiring ground-truth answers. ARES, in particular, offers statistical confidence bounds and modular component scoring, enabling accurate diagnostics even across domains.

End-to-end evaluation often obscures specific failure points within the RAG pipeline (such as suboptimal chunking, inaccurate retrieval, ineffective reranking, or hallucinated generation) making it difficult to isolate which component is responsible for performance degradation. To address this, eRAG~\cite{salemi2024evaluating} proposes a document-level relevance scoring based on LLM output, which correlates better with downstream QA performance than traditional query-document metrics. CoFE-RAG~\cite{liu2024cofe} expands this perspective by evaluating all stages: chunking, retrieval, reranking, and generation. These techniques allow developers to diagnose failure sources within the RAG pipeline and improve system interpretability. The RAGAS framework~\cite{es2023ragas} offers a comprehensive and modern evaluation pipeline, encompassing dataset generation from documents, LLM-based metrics, and a modular evaluation architecture. As it is the only actively maintained framework of its kind, we adopt RAGAS as the primary baseline for evaluating the performance of our proposed method. 
Even though all of these end-to-end evaluation frameworks have made significant strides, they rely on some degree of manual validation for QA dataset generation. Moreover, their metrics have notable limitations: rule-based metrics often fail to capture subtle semantic nuances, while works that use LLM-based metrics are rarely benchmarked for alignment with human judgment. As a result, such metrics may not fully reflect the intended evaluation objectives or agree with human assessments.

\subsection{Tuning LLM-as-a-Judge Evaluation}

Recent advances in LLM evaluation have increasingly focused on methods that improve alignment with human judgment and reduce variability in scoring. AutoCalibrate~\cite{liu2023autocalibrate} auto-tunes the evaluation prompts for better human alignment, ensuring that the scores reflect the actual preferences of the user. DSPy \cite{khattab2024dspy, khattab2022demonstrate} is a declarative framework that enables the programmatic creation and refinement of prompts for LLMs. Evaluation frameworks such as PoLL~\cite{verga2024poll} advocate a panel of various LLM evaluators to reduce bias and variance in generation scoring. These are powerful methods that have yet to be applied in RAG evaluation.


\subsection{Synthetic QA Data Generation}

The evaluation of RAG systems typically relies on a question-answering (QA) datasets. However, such QA datasets are often unavailable or insufficient in specialized domains~\cite{bai2024hard}. This limitation has driven a surge in research focused on generating high-quality synthetic QA data.

Alberti et al.~\cite{alberti2019synthetic} pioneered a round-trip consistency approach that combines answer extraction, question generation, and answer re-verification to create high-confidence Question-Answer-Context (QAC) triplets. Shakeri et al.~\cite{shakeri2020end} proposed an end-to-end transformer-based generator that outputs both the question and the answer from a given passage.

Recent QA data generation has increasingly shifted toward LLM-based approaches. Bai et al.~\cite{bai2024hard} tailored prompt engineering and summarization strategies to generate more challenging clinical QAC triplets from EHRs. Bohnet et al.~\cite{bohnet2024long} leveraged long-context LLMs to generate QAC triplets over entire books, demonstrating that automatic generation can rival or surpass human-crafted data sets in narrative domains.
RAGAS~\cite{es2023ragas} leverages agentic LLM designs and knowledge graph to generate context-rich and diverse question–answer pairs, with the LLMs simulating multiple personas during the generation process.

While these synthetic QA generation frameworks are impressive, none of them employ a fully automatic end-to-end pipeline that leverages LLM-based metrics to assess QA quality. RAGAS attempts to accomplish this yet their QA generation underperforms even on their own native metrics.

\section{Methodology} \label{methodology}

\begin{figure}[!t]
    \centering
    \includegraphics[width=3.4in]{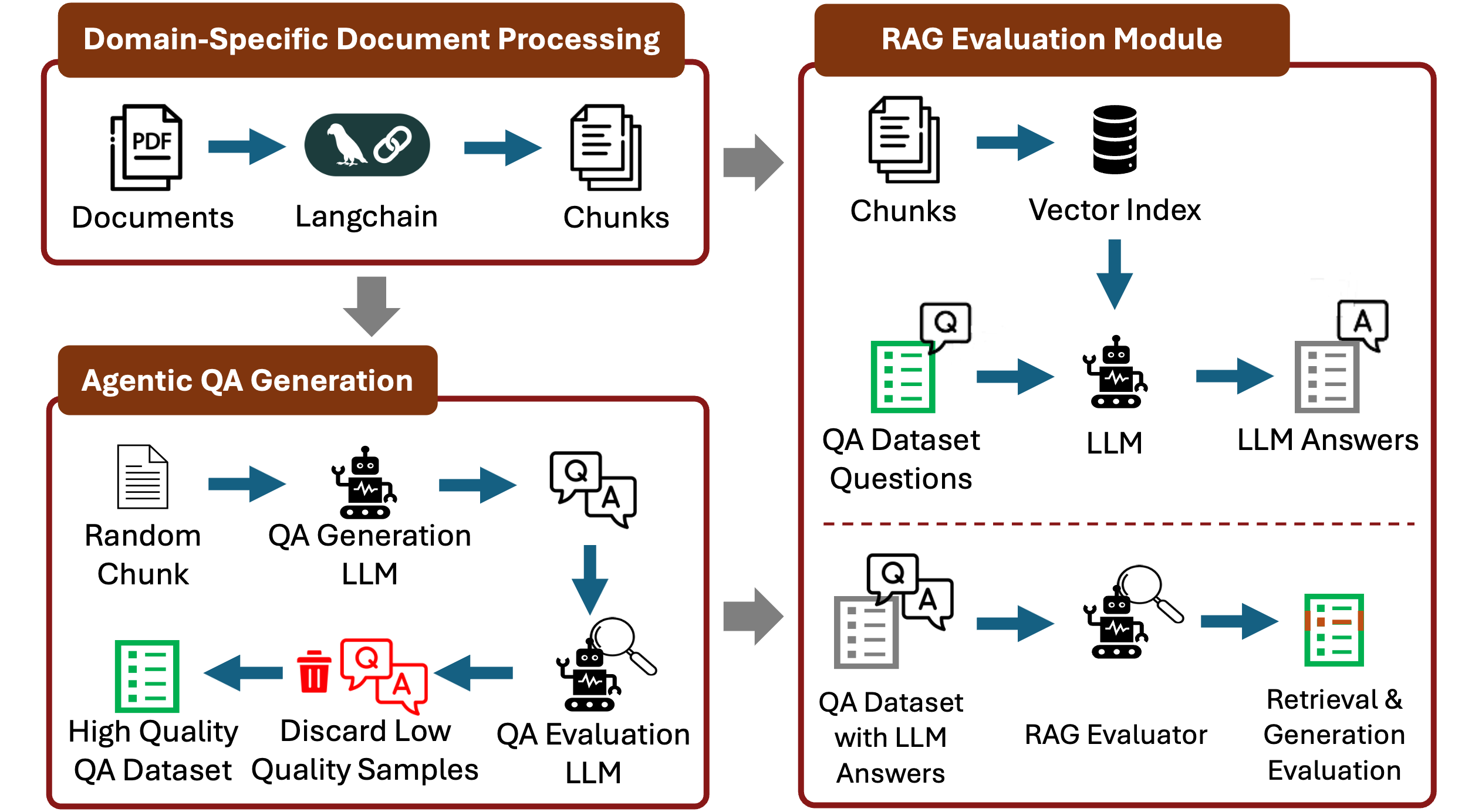}
    \caption{Overview of the RAGalyst framework that consists of three modules: a pre-processing module that transforms domain-specific documents into text chunks, a QA generation pipeline for producing synthetic question–answer-context datasets, and an evaluation module for assessing RAG system performance.}
    \label{fig:framework_overview}
\end{figure}

This section presents an agentic framework for evaluating RAG systems in domain-specific contexts. The framework includes a document preprocessing toolkit, agentic QA generation pipeline, and a set of LLM-based evaluation metrics. 
Figure~\ref{fig:framework_overview} illustrates the overall framework. 

\subsection{Domain-Specific Documents Preprocessing}

The preprocessing of domain-specific documents is the first and most critical step in this RAG framework, as it directly supports both the QA generation and evaluation stages. Without careful handling of document formatting and structure, the quality of retrieval and generated responses suffer significantly.

The framework first leverages \texttt{LangChain}\footnote{\url{https://www.langchain.com/}} to parse PDF, markdown, and plain text and then divides the documents into smaller units, called chunks, using a token-based chunking strategy. The size of these chunks plays a vital role in downstream performance. Chunks that are too small may lack context, leading to incomplete retrievals. On the other hand, chunks that are too large can dilute relevance and exceed the model context limits. Chunk size significantly impacts retrieval performance across domains due to differences in document structure and content. Optimal chunk size can vary by more than 20\% depending on the domain, influencing both precision and recall metrics in RAG evaluations~\cite{jadon2025enhancing}. Since there is no standardized text chunking strategy, we follow OpenAI's file search tool.\footnote{\url{https://platform.openai.com/docs/assistants/tools/file-search}} and select a default of a maximum 800 tokens per chunk with an overlap between chunks of 400 tokens. Afterwards, the chunks are vectorized with the selected embedding model and are stored in a vector database.

\subsection{Agentic QA Generation Pipeline}

A critical requirement for evaluating RAG systems is the availability of high-quality annotated datasets. These datasets must contain: 
\begin{enumerate}
    \item Domain-relevant questions.
    \item Ground-truth answers.
    \item Context used to answer each question.
\end{enumerate} 

However, for many classified or sensitive domains, such annotated datasets are non-existent. Manual annotation of such datasets is often impractical due to confidentiality restrictions, inconsistent formatting, and the high cost of human labeling.

To address this gap, the agentic QA generation pipeline leverages LLMs in a role-driven, autonomous fashion to emulate users to produce context-based, high-fidelity question-answer pairs based on preprocessed document chunks. Each generated answer is grounded directly in its source context and quality is validated using multiple LLM-based metrics from the evaluation module (more details in~\ref{evaluation_module}) to ensure alignment and quality. This approach eliminates the need for labor-intensive annotation while ensuring consistency, reproducibility, and scalability across various domains.

The pipeline operates in three main steps:

\paragraph{Context Sampling} 
Chunks generated during the document chunking step are randomly sampled and used as reference contexts for question, answer, context triplet (QAC).

\paragraph{Prompting and QA Generation}
An agentic LLM assumes the role of a user to generate a question that is related to and answerable by the sampled context. The questions is evaluated to ensure it is specific and unambiguous. Then, another agent assumes the role of a subject matter expert to answer the question to generate the ground truth answer.
    
\paragraph{Validation and Filtering}
To ensure quality, the LLM-based evaluation module validates the generated QA pairs using Answerability, Faithfulness and Answer Relevance metrics (more information in Section~\ref{metrics}). If the sample is unable to meet the thresholds, it is discarded to preserve dataset integrity. These thresholds are hyperparameters that control quality strictness. The higher values enforce stronger filtering but result in longer generation times due to an increased number of candidate QAs being discarded.

\begin{table*}[!t]
\caption{Summary of Evaluation Metrics for Dataset Generation and RAG Evaluation}
\label{tab:rag_metrics_summary}
\centering
\begin{tabular}{>{\raggedright\arraybackslash}p{3cm} >{\raggedright\arraybackslash}p{4.5cm} c >{\raggedright\arraybackslash}p{3.5cm} c}
\toprule
\textbf{Metric Name} & \textbf{Use} & \textbf{Scale} & \textbf{Used In} & \textbf{Framework} \\
\midrule
Answer Correctness   & Evaluates how accurately the generated answer matches the ground truth & 0--1.0 & RAG Evaluation & Ours \\
Answerability         & Evaluates whether the question can be answered using only the provided context & 0 or 1 & Dataset Generation & Ours \\
Faithfulness         & Measures how well the answer is grounded in the provided context & 0--1.0 & Dataset Generation, RAG Evaluation & RAGAS \\
Answer Relevance     & Assesses whether the answer directly and meaningfully addresses the question & 0--1.0 & Dataset Generation, RAG Evaluation & RAGAS \\
Recall@K             & Measures whether the ground truth context appears in the top-$K$ retrieved documents & 0 or 1 & Retrieval Evaluation & Standard \\
MRR                  & Computes the inverse rank of the first correct retrieval to evaluate re-ranking quality & 0--1.0 & Retrieval Evaluation & Standard \\
\bottomrule
\end{tabular}
\end{table*}

\subsection{RAG Evaluation Module}\label{evaluation_module}

The RAG evaluation module leverages LLM-as-a-judge to enable automated, scalable, and consistent scoring across diverse settings. Additionally, the framework also provides essential heuristic metrics for evaluating retrieval performance.

\subsubsection{Retrieval Evaluation}
Mean Reciprocal Rank (MRR) and Recall@K are two standard information retrieval metrics used to quantify retrieval effectiveness \cite{yu2024evaluation}.

MRR computes the average of the reciprocal ranks of the first relevant document across queries, effectively measuring how early the correct context appears in the retrieval list. MRR is particularly well-suited for assessing re-ranking strategies in RAG systems, as it emphasizes placing relevant information as close to the top of the ranked list as possible.

\noindent MRR is defined as:

\[
\text{MRR} = \frac{1}{|Q|} \sum_{i=1}^{|Q|} \frac{1}{\text{rank}_i}
\]

\noindent where:
\begin{itemize}
    \item $|Q|$ is the total number of queries.
    \item $\text{rank}_i$ is the position of the ground truth context in the ranked list for the $i$-th query.
\end{itemize}

Recall@K, also referred to as Hit Rate@K, is the average recall across multiple queries where ground truth context appears within the top-$k$ retrieved results for a given query. For each query, recall is a binary metric. It returns 1 if the ground truth context is found within the top $k$ or 0 if not. This metric is particularly important for RAG systems since it calculates the chance the ground truth context is included among the retrieved top candidates.

Recall@$k$ is defined as:

\[
\text{Recall@K} = \frac{1}{|Q|} \sum_{i=1}^{|Q|} \mathbf{1}(\text{rank}_i \leq k)
\]

\noindent where:
\begin{itemize}
    \item $|Q|$ is the total number of queries.
    \item $\text{rank}_i$ is the position of the ground truth document for the $i$-th query.
    \item $\mathbf{1}(\cdot)$ is the indicator function, which returns 1 if the condition inside is true, and 0 otherwise.
    \item $k$ is the cutoff rank threshold (for example, Recall@3 considers the top 3 results).
\end{itemize}

\subsubsection{LLM-Based Evaluation}\label{metrics}

To ensure the quality of the synthetic QA dataset and support downstream RAG evaluation, the framework combines established metrics with a novel LLM-based evaluation approach. While standard metrics from the RAGAS framework~\cite{es2023ragas} are effective for detecting hallucinations and assessing answer relevance, they fall short in evaluating the quality of generated questions and the correctness of RAG-generated answers. To address these limitations, the framework introduces custom LLM-as-a-judge metrics, which leverage language models to better account for linguistic variation, paraphrasing, and latent knowledge~\cite{Gu2025Survey, Gao2023GEval, Zheng2023Judge, Kocmi2023LLMEval}. This LLM-based strategy enables a more robust, human-aligned evaluation pipeline. The key metrics introduced are:

\begin{itemize}
\item \textbf{Answer Correctness}: A custom LLM-as-a-judge metric that compares a generated answer to the reference (ground truth) answer. The LLM scores semantic alignment on a continuous scale, offering both a numerical score and a rationale. This enables flexible evaluation even when surface-level phrasing differs.

\item \textbf{Answerability}: Assesses whether a generated question is fully supported by the retrieved context, without relying on external knowledge. This enforces quality in our synthetic QA dataset generation.
\end{itemize}

Table~\ref{tab:rag_metrics_summary} summarizes the definitions of all metrics and their roles within our framework.
\section{Framework Validation}\label{framework_validation}

This section validates the Agentic Framework for Domain-Specific RAG Evaluation (RAGalyst) by assessing its accuracy, reliability, and practical suitability for domain-specific document tasks. The primary objective is to determine whether the framework’s evaluation metrics and data generation methods are effective enough to support real-world use cases.

First, we assess the reliability of the LLM-as-a-judge evaluation metrics Answer Correctness and Answerability—along with prompt-optimized variants. These metrics are tested against human-annotated references to ensure alignment with human judgment. 

Second, the synthetic QA datasets generated by the agentic QA generation pipeline are evaluated against publicly available domain-specific QA sets, and datasets generated by RAGAS, to assess its fidelity and reliability in evaluating RAG systems. This comparison aims to verify whether the synthetic data produced by this framework serves as a viable substitute for manually curated datasets in high-stakes domains.

\subsection{Metrics Validation}

We evaluate the performance of both Answer Correctness and Answerability by computing the Spearman correlation between their scores and human annotations. To calculate the standard error of a Spearman correlation, we use Bonnett and Wright \cite{bonett2000sample} standard error (SE) approximation:
\[
\text{SE}(\rho_s) = \sqrt{\frac{1+\frac{\rho_s^2}{2}}{n-3}} 
\]
Where:
\begin{itemize}
    \item $\text{SE}(\rho_s)$ is the standard error of the Spearman correlation.
    \item $\rho_s$ is the expected Spearman correlation coefficient.
    \item $n$ is the sample size.
\end{itemize}

Since a manually crafted Answer Correctness and Answerability prompt is unlikely to be optimal, we turn to DSPy’s \cite{khattab2022demonstrate, khattab2024dspy} COPRO, MIPROv2, and LabeledFewShot optimizers. These optimizers systematically refine prompts using supervision and feedback signals, improving alignment with human annotations.

COPRO and MIPROv2 are automatic instruction optimizers where the instructions in the prompt are tweaked and validated. The COPRO optimizer generates and refines new instructions for each step, and optimizes them with coordinate ascent on a defined metric function. MIPROv2 generates instructions and few-shot examples in each step of optimization. The instruction generation is data-aware and demonstration-aware, and uses Bayesian Optimization to effectively search over the space of generation instructions/demonstrations. 

LabeledFewShot is a automatic few-shot learning optimizer. It randomly samples $k$ examples from a labeled dataset and adds them to the prompt as demonstrations.

\begin{table*}[!t]
    \caption{Human Alignment of Methods for Answer Correctness and Answerability }
    \label{tab:ac_validation}
    \centering
    \setlength{\tabcolsep}{5pt} 
    \begin{tabular}{llcc}
        \toprule
        \textbf{Method} & \textbf{Model} & \textbf{Answer Correctness} $\boldsymbol{\rho_s}$ & \textbf{Answerability} $\boldsymbol{\rho_s}$ \\
        \midrule
        Cosine Similarity & Qwen3-Embedding-8B & 0.622 & -- \\
        RAGAS & gpt-4o-mini & 0.836 & -- \\
        \midrule
        Ours      &   gemma-3-27b-it & 0.862 & 0.596 \\
        Ours & Qwen3-30B-A3B-Instruct-2507 & 0.851 & 0.605 \\
        Ours & gemini-2.5-flash-lite & 0.849 & 0.665 \\
        Ours & gemini-2.5-flash & 0.777 & 0.749 \\
        Ours & gemini-2.5-pro & 0.805 & \textbf{0.752} \\
        Ours & gpt-4o-mini & \textbf{0.874} & 0.700 \\
        Ours & gpt-4.1-nano & 0.857 & 0.436 \\
        Ours & gpt-4.1-mini & 0.873 & 0.670 \\
        Ours & gpt-4.1 & 0.857 & 0.723 \\
        \midrule
        Ours: COPRO Optimized & gpt-4o-mini &  0.881 & \textbf{0.670} \\
        Ours: MIPROv2 Optimized & gpt-4o-mini  & 0.887 & 0.669 \\
        Ours: MIPROv2 \& LabeledFewShot Optimized & gpt-4o-mini & \textbf{0.894} & -- \\
        Ours: LabeledFewShot Optimized & gpt-4o-mini & -- & 0.632 \\
        \bottomrule
    \end{tabular}
\end{table*}

\subsubsection{Answer Correctness}\label{sec:ac_validation}
This metric was benchmarked against the Semantic Textual Similarity Benchmark (STS-B), part of the GLUE benchmark suite\footnote{\url{https://gluebenchmark.com/tasks/}}. STS-B consists of sentence pairs drawn from sources such as news headlines, image captions, and online forums, with each pair annotated by humans with a similarity score ranging from 0 (completely dissimilar) to 5 (semantically equivalent). To adapt the dataset to the evaluation task, the first sentence in each pair was treated as the ground-truth answer and the second as the model-generated answer. The original STS-B scores were normalized from a range of 0–5 to a range of 0.0–1.0 to align with the output of our LLM-based Answer Correctness metric.

In this validation, we compare our proposed Answer Correctness metric—along with a prompt-optimized variant—against baseline methods including cosine similarity and RAGAS's Answer Correctness. Both RAGAS and our framework utilize the GPT-4o-mini model at a temperature of 0 to ensure a consistent evaluation backbone across approaches, but we also compare against other state of the art LLMs. Cosine similarity is computed using the Qwen3-Embedding-8B model, the top-ranked model on the MTEB leaderboard~\footnote{\url{https://huggingface.co/spaces/mteb/leaderboard}} for the STS task at the time of writing.

On a test set of 500 randomly sampled STS-B sentence pairs, our Answer Correctness metric achieves a mean Spearman correlation of 0.874 with a standard error of 0.053 with GPT-4o-mini. This outperforms both Cosine Similarity ($\rho_s$ = 0.622, SE = 0.049) and RAGAS ($\rho_s$ = 0.843, SE = 0.052), demonstrating stronger alignment with human-annotated similarity scores. 

We first refine the prompt instructions using both the COPRO and MIPROv2 optimizers on a training set of 500 STS-B samples, and evaluate performance on the test set. We adopt DSPy's default parameters for both optimizers and use GPT-4o-mini to generate optimized prompt candidates. The metric achieves $\rho_s$ = 0.881 (SE = 0.053) with COPRO and $\rho_s$ = 0.887 (SE = 0.053) with MIPROv2.

Since the MIPROv2 optimized Answer Correctness metric performs the best, we further optimize its prompt with LabeledFewShot optimizer which pushes performance to $\rho_s$ = 0.894 (SE = 0.053) at a $k$ of 8. We ablate $k$ in Figure \ref{fig:prompt_optimization_ablation}. This combination of the MIPROv2 and LabeledFewShot optimizers performs the best for the Answer Correctness metric, and therefore use this version of the metric throughout the remainder of this paper.

\subsubsection{Answerability}

The Answerability metric was validated using the Stanford Question Answering Dataset 2.0 (SQuAD 2.0)\footnote{\url{https://rajpurkar.github.io/SQuAD-explorer/}}. SQuAD 2.0 extends the original SQuAD dataset by including over 50,000 unanswerable questions written adversarially to resemble answerable ones. Each entry in the dataset contains a question, a context passage, a ground-truth answer, and a binary flag indicating whether the question is answerable given the context. This makes it an ideal benchmark for evaluating the Answerability of QACs, as it directly tests whether the provided context alone is sufficient to support a meaningful response. Each LLM evaluates the Answerability of the QACs with a temperature of 0.

Using Gemini 2.5 Pro on a test set of 500 randomly sampled SQuAD 2.0 QA pairs, our Answerability metric achieves a mean Spearman correlation of 0.752 with a standard error of 0.051

We first refine the prompt instructions using both the COPRO and MIPROv2 optimizers on a training set of 500 SQuAD 2.0 QA pairs, and evaluate performance on a the test set. We adopt DSPy's default parameters for both optimizers and use GPT-4o-mini to generate optimized prompt candidates. The metric achieves $\rho_s$ = 0.670 (SE = 0.050) with COPRO and $\rho_s$ = 0.669 (SE = 0.050) with MIRPOv2. 

\begin{figure}[!t]
    \centering
    \includegraphics[width=3.4in]{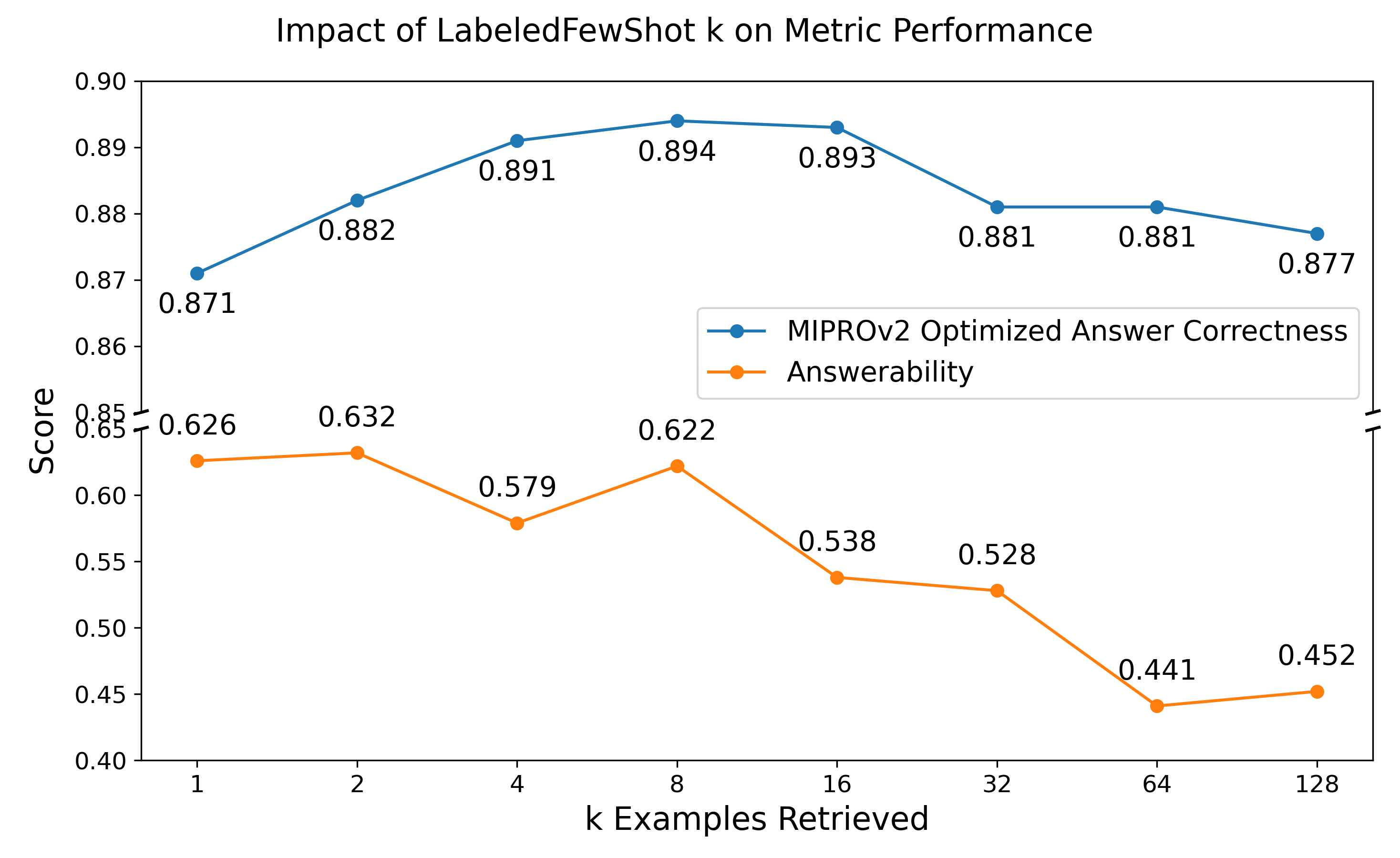}
    \caption{We ablate both the MIPROv2-optimized Answer Correctness metric and the non-optimized Answerability metric using the LabeledFewShot optimizer. Our results show that Answer Correctness achieves its best performance with 8 examples, whereas LabeledFewShot optimization provides no improvement over our handcrafted prompt for Answerability.}
    \label{fig:prompt_optimization_ablation}
\end{figure}

Since neither automatic instruction optimizers improve performance, we apply LabeledFewShot optimizer on the non-optimized Answerability metric with GPT-4o-mini. This achieves $\rho_s$ = 0.632 (SE = 0.049) at a a $k$ of 2. We ablate $k$ in Figure \ref{fig:prompt_optimization_ablation}. All optimizers do not improve performance for the Answerability metric, and therefore use the non-prompt optimized version of Answerability throughout the remainder of this paper.

\subsubsection{Metric's LLM Selection}

The results in Table \ref{tab:ac_validation} suggest that GPT-4o-mini provides the best balance of performance metrics, inference speed, and cost. For this reason, \textbf{GPT-4o-mini is used for all LLM-based metrics throughout the remainder of this paper.}

\subsection{Agentic QA Dataset Generation Pipeline Validation}\label{dataset_eval}

To rigorously validate the performance of our Agentic QA generation pipeline, we constructed three domain-specific datasets using our framework. The domains military operations (army), cybersecurity, and bridge engineering (engineering) were selected for their real-world importance and the abundance of diverse, unstructured source material such as technical documents, reference books, and cheatsheets. For each domain, we generated 500 QA pairs, resulting in a total of 1,500 samples that reflect a broad spectrum of question types and contextual difficulty levels.

To establish a fair and consistent baseline for comparison, we used the same source materials to generate three additional datasets using the RAGAS framework. Both frameworks were configured to produce Single Hop-specific QA datasets, which require no or minimal reasoning. This controlled setup ensures that the only variable under evaluation is the generation method itself. We evaluated both approaches using a mixture of our proposed metrics (Answerability) and the RAGAS metrics (Faithfulness, Answer Relevance), allowing us to assess our performance against a strong existing baseline.

In addition to synthetic comparisons, we incorporated a benchmark evaluation against human-annotated datasets. These datasets were selected based on their widespread use in the QA and RAG research communities, their utility in real-world applications, and their sample sizes being large enough to ensure statistical significance. 

\begin{itemize}
    \item The COVID-QA dataset from deepset comprises of 2,019 expert-annotated QA pairs drawn from 147 COVID‑19 scientific articles by 15 biomedical specialists~\cite{moeller2020covidqa}. It is highly relevant to real-world biomedical information retrieval tasks.
    
    \item The RepLiQA dataset by ServiceNow Research consists of approximately 90,000 question–answer–context triplets, all carefully written and annotated by human experts using fictional yet coherent narratives~\cite{pal2024replica}. It is specifically designed to rigorously evaluate a model’s ability to reason over unseen, non-factual content.
\end{itemize}

By applying the same evaluation framework to both the machine-generated and human-annotated datasets, we are able to observe how closely our pipeline aligns with human-created content across different domains. This evaluation approach supports a more balanced assessment of the method's reliability and generalizability.

Table~\ref{tab:ours_ragas} demonstrates that our framework consistently outperforms RAGAS across both its native metrics and our evaluation criteria. This performance gap stems primarily from two critical limitations in the RAGAS pipeline. First, RAGAS lacks an effective filtering mechanism to exclude low-quality samples, leading to noisier datasets. Second, its use of a single multitask prompt for generating both questions and answers reduces generation quality, aligning with prior findings that multitasking degrades large language model performance~\cite{gozzi2024comparative}. Notably, while the RAGAS-generated dataset achieves reasonable scores in Faithfulness and Answer Relevance, the dataset's Answerability remains significantly lower. This pattern suggests that the model generates answers that appear contextually appropriate and responsive but are not explicitly supported by the retrieved evidence, indicating a tendency toward informed hallucination rather than faithful grounding.

Although our custom QA generation pipeline produces higher quality samples, it operates at a slower rate. It generates 100 QA samples at an average of 0.881 samples per minute (single-threaded) and 7.039 spm (16-threaded). In contrast, RAGAS achieves 14.627 spm (single-threaded) and 33.842 spm (16-threaded). This slower generation rate is primarily due to our more stringent QA filtering.

We provide an example generated QA from the Military Operations domain.

\begin{tcolorbox}[colback=gray!5!white, colframe=gray!60!black, title={Example Generated QA}]

\textbf{Text Chunk as Context:}

\small
\begin{quote}
Chapter 7 \\
7-6 TC 3-21.76 26 April 2017 \\
7-29. R\&S teams move using a technique such as the cloverleaf method to move to successive OPs. (See figure 7-1.) In this method, R\&S teams avoid paralleling the objective site, maintain extreme stealth, do not cross the LOA, and maximize the use of available cover and concealment. \\
7-30. During the conduct of the reconnaissance, each R\&S team returns to the RP when any of the following occurs: \\
\hspace*{1em}-- All their PIR is gathered. \\
\hspace*{1em}-- The LOA is reached. \\
\hspace*{1em}-- The allocated time to conduct the reconnaissance has elapsed. \\
\hspace*{1em}-- Enemy contact is made. \\[2pt]
\textit{LEGEND:} ORP – objective rally point; RP – release point; S\&O – surveillance and observation.
\end{quote}

\normalsize
\medskip
\textbf{Question:} What action must an R\&S team take if they make enemy contact during reconnaissance?

\medskip
\textbf{Answer:} If an R\&S team makes enemy contact during reconnaissance, they must return to the RP.

\end{tcolorbox}

\begin{table}
    \centering
    \caption{Synthetic QA Dataset Generation Performance Across Domains}
    \begin{tabular}{llccc}
        \toprule
        \textbf{Domain} & \textbf{Metric} & \textbf{Human} & \textbf{Ours} & \textbf{RAGAS} \\
        \midrule
        \multirow{3}{*}{COVID-QA}
            & Faithfulness      & 0.894 & \textbf{0.989} & 0.917 \\
            & Answerability     & 0.620 & \textbf{0.994} & 0.418 \\
            & Answer Relevance  & 0.399 & \textbf{0.947} & 0.748 \\
        \midrule
        \multirow{3}{*}{RepLIQA}
            & Faithfulness      & 0.774 & \textbf{0.994} & 0.957 \\
            & Answerability     & 0.740 & \textbf{0.998} & 0.916 \\
            & Answer Relevance  & 0.475 & \textbf{0.962} & 0.770 \\
        \midrule
        \multirow{3}{*}{Army} 
            & Faithfulness     & -- & \textbf{0.977} & 0.868 \\
            & Answerability    & -- & \textbf{0.994} & 0.618 \\
            & Answer Relevance & -- & \textbf{0.957} & 0.830 \\
        \midrule
        \multirow{3}{*}{Cybersecurity} 
            & Faithfulness     & -- & \textbf{0.991} & 0.797 \\
            & Answerability    & -- & \textbf{0.974} & 0.306 \\
            & Answer Relevance & -- & \textbf{0.961} & 0.783 \\
        \midrule
        \multirow{3}{*}{Engineering} 
            & Faithfulness     & -- & \textbf{0.988} & 0.839 \\
            & Answerability    & -- & \textbf{0.990} & 0.473 \\
            & Answer Relevance & -- & \textbf{0.958} & 0.797 \\
        \bottomrule
    \end{tabular}
    \label{tab:ours_ragas}
\end{table}
\section{Experiments}\label{experiments}

To demonstrate the applicability of RAGalyst, this section showcases the experimental evaluation of various RAG approaches using the framework. These experiments assess key components of RAG systems, including the embedding retrieval, LLM generation, and context length. In addition, we analyze the potential bias in LLM's preference for self-generated text, and we provide an analysis on reasons for low Answer Correctness.

\subsection{Embedding Retrieval Evaluation Across Domains}

\begin{figure}[!t]
    \centering
    \includegraphics[width=3.4in]{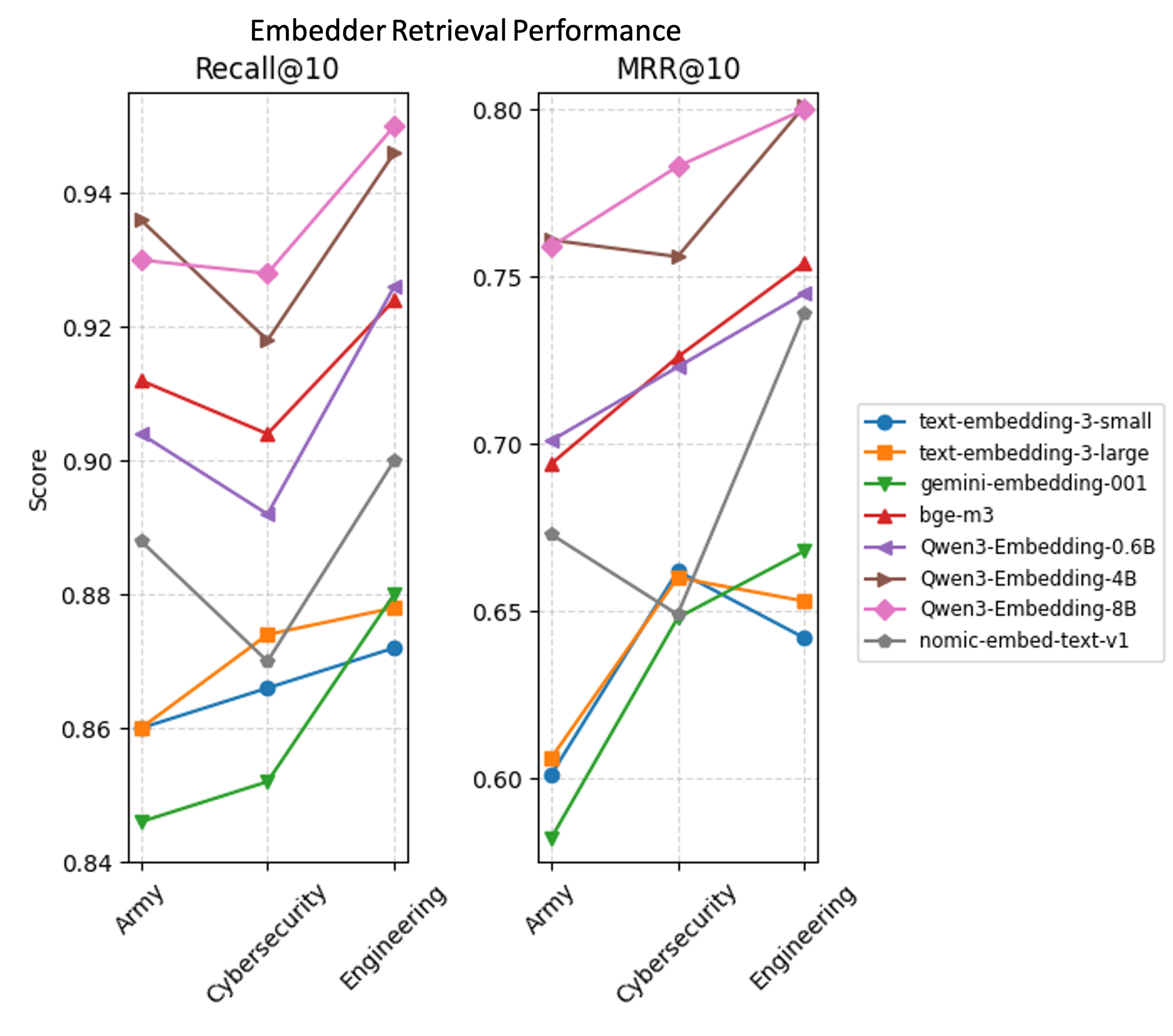}
    \caption{We evaluate retrieval with Recall@10 and MRR@10 metrics on a variety of embedding models on three different domains.}
    \label{fig:embedder_retrieval_eval}
\end{figure}

The performance of any RAG system is fundamentally dependent on the quality of its embedding model. The embedding model determines how well the system can semantically match queries with relevant document chunks. Poor retrieval due to weak embeddings will degrade the overall output. Therefore, selecting the right model is a critical design decision. In this hypothetical scenario, we will evaluate the retrieval performance of a diverse selection of embedding models. Our analysis will include models that rank highly on the MTEB leaderboard \cite{muennighoff2022mteb}, as well as a variety of open-source and closed-source models. We will also consider models of different sizes to assess the impact of model scale on performance.

To evaluate retrieval performance, we employ Recall@K and MRR@K on the datasets generated by our pipeline introduced in Section~\ref{dataset_eval}, using $k = 10$. Each document is chunked with a maximum size of 800 tokens and an overlap of 400 tokens between consecutive chunks. As shown in Figure~\ref{fig:embedder_retrieval_eval}, the results reveal the variation in embedding model performance across different domains. The Qwen3 family of models \cite{zhang2025qwen3} consistently performs well in retrieval tasks across various domains, reaffirming their high placement on the MTEB leaderboard. Despite its third-place ranking on MTEB, the gemini-embedding-001 \cite{lee2025gemini} model underperforms compared to other embedding models on these specific domains. Meanwhile, the BGE-M3 \cite{chen2024m3} embedding model shows performance comparable to the Qwen3-Embedding-0.6B model, even though its MTEB ranking is significantly lower. Notably, open-source models generally outperform their closed-source counterparts. An interesting observation within the Qwen3 family is that the smaller 4B model performs similarly to, and in some domains even surpasses, its larger 8B counterpart.

Performance across different domains is inconsistent for most embedding models. For example, recall scores for most models are weaker in the cybersecurity domain compared to other areas. However, the text-embedding-3 family of embedding models defies this trend, performing much better in cybersecurity relative to the other domains. This suggests they were likely trained on a larger volume of cybersecurity-specific data. A similar pattern is observed with mean reciprocal rank (MRR), underscoring that embedding model effectiveness is highly domain-dependent. This inconsistency reinforces the need to evaluate RAG systems with domain-specific benchmarks.

\subsection{Domain Specific LLM Generation Evaluation}
\label{sec:llm_with_rag}

This experiment evaluates the domain specific RAG performance of a diverse set of LLMs on Answer Correctness, Faithfulness, and Answer Relevancy. To ensure consistency, each model receives the same set of 10 retrieved context chunks with a maximum chunks size of 800 tokens, and chunk overlap of 400 tokens. For retrieval, we use the top-performing model on the MTEB retrieval task, Qwen3-Embedding-8B.

\begin{figure}[!t]
    \centering
    \includegraphics[width=3.4in]{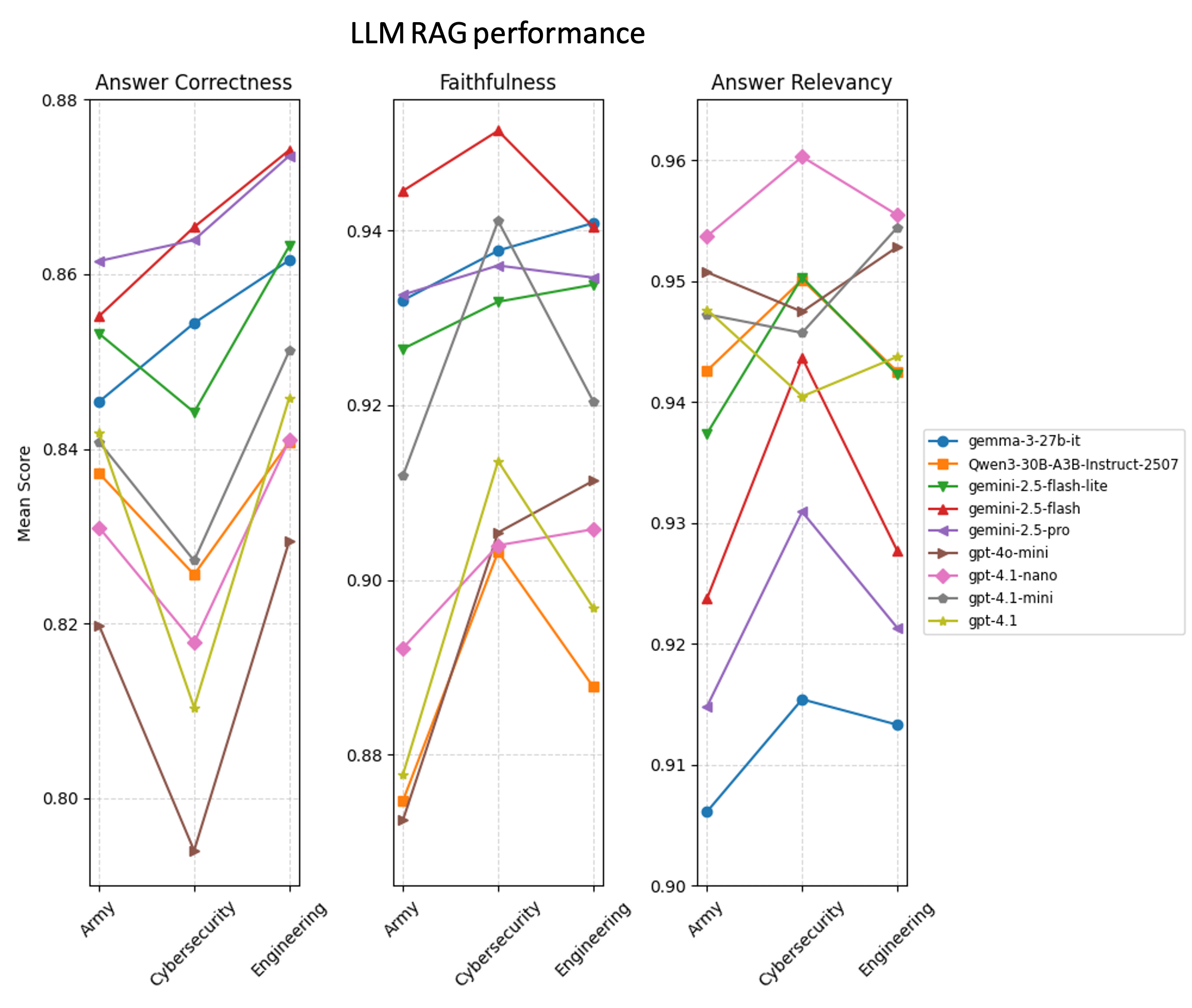}
    \caption{We evaluate LLM generation with the Answer Correctness, Faithfulness, and Answer Relevancy metrics on three different domains.}
    \label{fig:llm_with_rag_eval}
\end{figure}

As shown in Figure \ref{fig:llm_with_rag_eval}, Gemini-2.5-flash \cite{comanici2025gemini} shows the strongest overall performance in Answer Correctness, achieving the highest scores in cybersecurity and bridge engineering domains. Most models struggle with the cybersecurity domain, with several models (including GPT-4o-mini, GPT-4.1, and Qwen3) dropping significantly in this domain compared to army and bridge engineering domains. This may be due to the drop in recall for Qwen3-Embeddding-8B embedding model, as seen in Figure \ref{fig:embedder_retrieval_eval}. Additionally, the Google family of models (Gemini and Gemma3 \cite{team2025gemma}) perform better than other models.

Gemini-2.5-flash demonstrates the strongest overall Faithfulness, achieving top scores in the army and cybersecurity and remaining competitive in bridge engineering domains. Models generally perform better in cybersecurity on Faithfulness, suggesting that the LLMs may not know the answer to the question from parametric knowledge and are leaning more heavily on the retrieved context chunks.

GPT-4.1-nano performs the best in all domains in Answer Relevancy. In general, the GPT family of models score higher than the Gemini models in this metric. This could be due to the fact that GPT models are more wordy in their responses, making it more likely for the responses to be more relevant to the question.

No single model dominates all 3 metrics in all 3 domains. There doesn't seem to be a significant advantage of closed-sourced LLMs compared to open-sourced ones (Qwen3 and Gemma3) in these domains. Additionally,  larger model sizes do not equate to better performance. This further highlighting the variability in performance across LLMs for different domains and the need for RAGalyst.

\subsection{Domain Specific Retrieved Chunks Evaluation}

To study the sensitivity of evaluation metrics to the number of retrieved chunks, we varied the number of chunks from 1 to 10 using Gemma3-4B. Across the three domains, the optimal number of chunks for maximizing Answer Correctness differed, suggesting that retrieval depth should be tuned to the domain. Moreover, all metrics exhibited substantial variation across domains, underscoring the importance of domain-specific RAG evaluation rather than a one-size-fits-all approach.

Despite these differences, some consistent trends emerge across metrics. Faithfulness tends to decline slightly as the number of chunks increases, likely because the LLM must contend with more irrelevant information. In contrast, Answer Relevancy generally improves with additional chunks, as the broader context encourages the model to at least address the question. Answer Correctness shows a peaked behavior, typically highest when 3–5 chunks are retrieved. With too few chunks, relevant information is often missed, reducing correctness. With too many, the relevant chunk risks being diluted by irrelevant ones, leading to distraction and lower performance.

\begin{figure}[!t]
    \centering
    \includegraphics[width=3.4in]{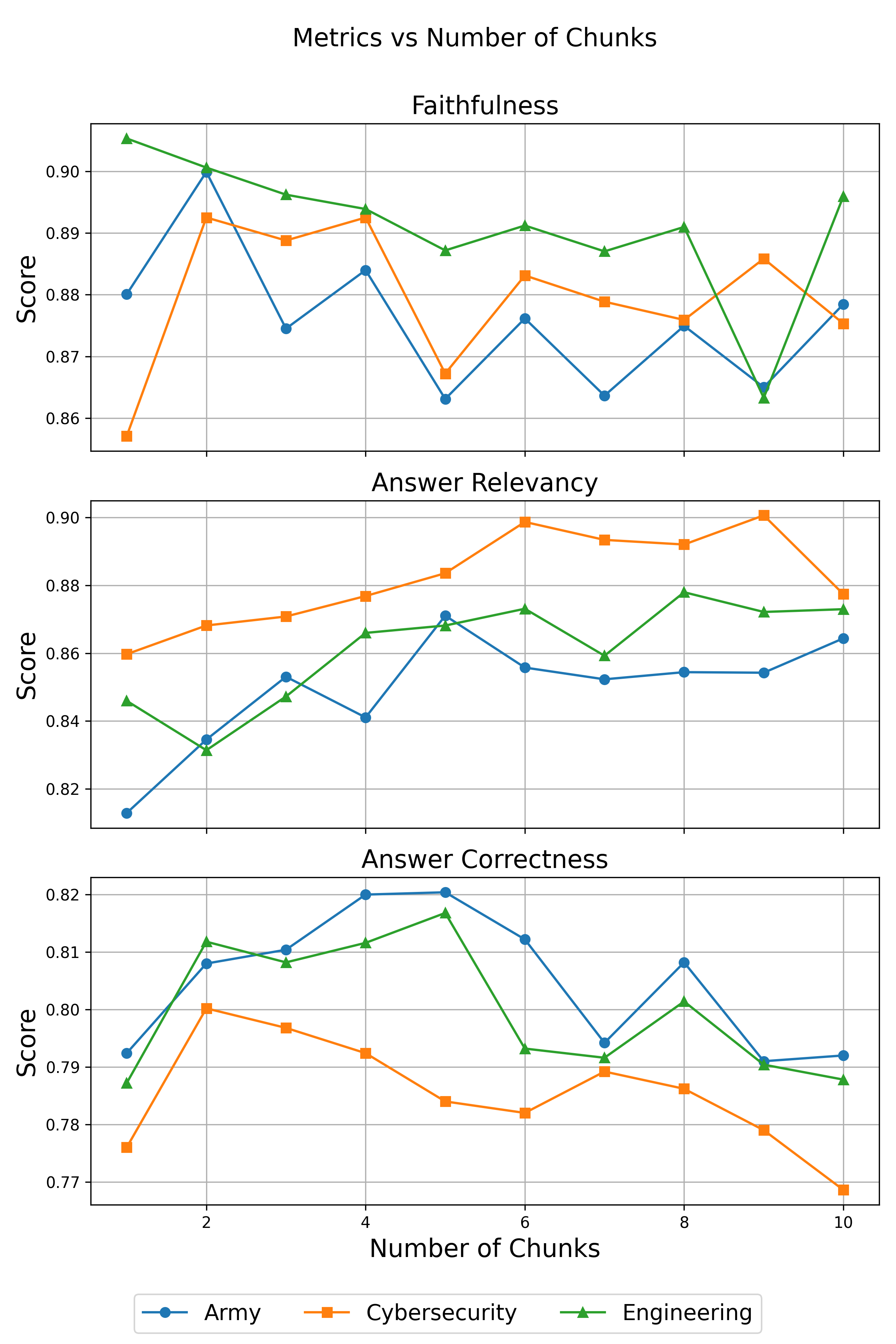}
    \caption{We ablate the number of chunks retrieved with Gemma3-4B to assess the effect on LLM generation performance on Answer Correctness, Faithfulness and Answer Relevancy. This figure shows the each metric responds differently to the number of chunks retrieved, and that the ideal number of chunks retrieved to maximize Answer Correctness will vary.}
    \label{fig:chunks_ablation}
\end{figure}

\subsection{Low Answer Correctness Analysis}

\begin{table}[htbp]
\centering
\caption{Low Correctness Reasons Counts as a Percentage of Total QAs Across Three Domains (Multiple Reasons per QA Possible)}
\label{tab:failure_counts_grouped}
\small
\begin{tabular}{lrrrr}
\toprule
\multicolumn{5}{c}{\textbf{All QAs}} \\
\midrule
\textbf{Failure Types} & \textbf{Army} & \textbf{Cyber.} & \textbf{Engine.} & \textbf{Total} \\
Number of QAs                & 500 & 500 & 500 & 1500 \\
\midrule
\multicolumn{5}{l}{\textbf{RAG Failures}} \\
Over-Specificity       & 58.4\% & 81.0\% & 74.6\% & 71.3\% \\
Not Extracted          & 11.0\% & 13.0\% & 15.0\% & 13.0\% \\
Under-Specificity      & 9.8\% & 10.2\% & 10.4\% & 10.13\% \\
Missed Top Ranked      & 5.0\% & 6.6\% & 5.4\% & 5.7\% \\
Wrong Format           & 2.2\% & 0.6\% & 1.8\% & 1.5\% \\
\multicolumn{5}{l}{\textbf{LLM Failures}} \\
Context Inconsistency  & 6.6\% & 9.6\% & 8.8\% & 8.3\% \\
Factual Fabrication    & 2.6\% & 7.4\% & 8.4\% & 6.1\% \\
Factual Contradiction  & 2.6\% & 5.6\% & 3.6\% & 3.9\% \\
Logical Inconsistency  & 0.8\% & 1.0\% & 0.4\% & 0.7\% \\
Instruction Inconsistency & 0.4\% & 0.0\% & 0.0\% & 0.1\% \\
\textbf{No Failures}            & 31.1\% & 10.6\% & 14.4\% & 18.7\% \\
\bottomrule
\end{tabular}
\end{table}

\begin{figure}[!t]
    \centering
    \includegraphics[width=3.4in]{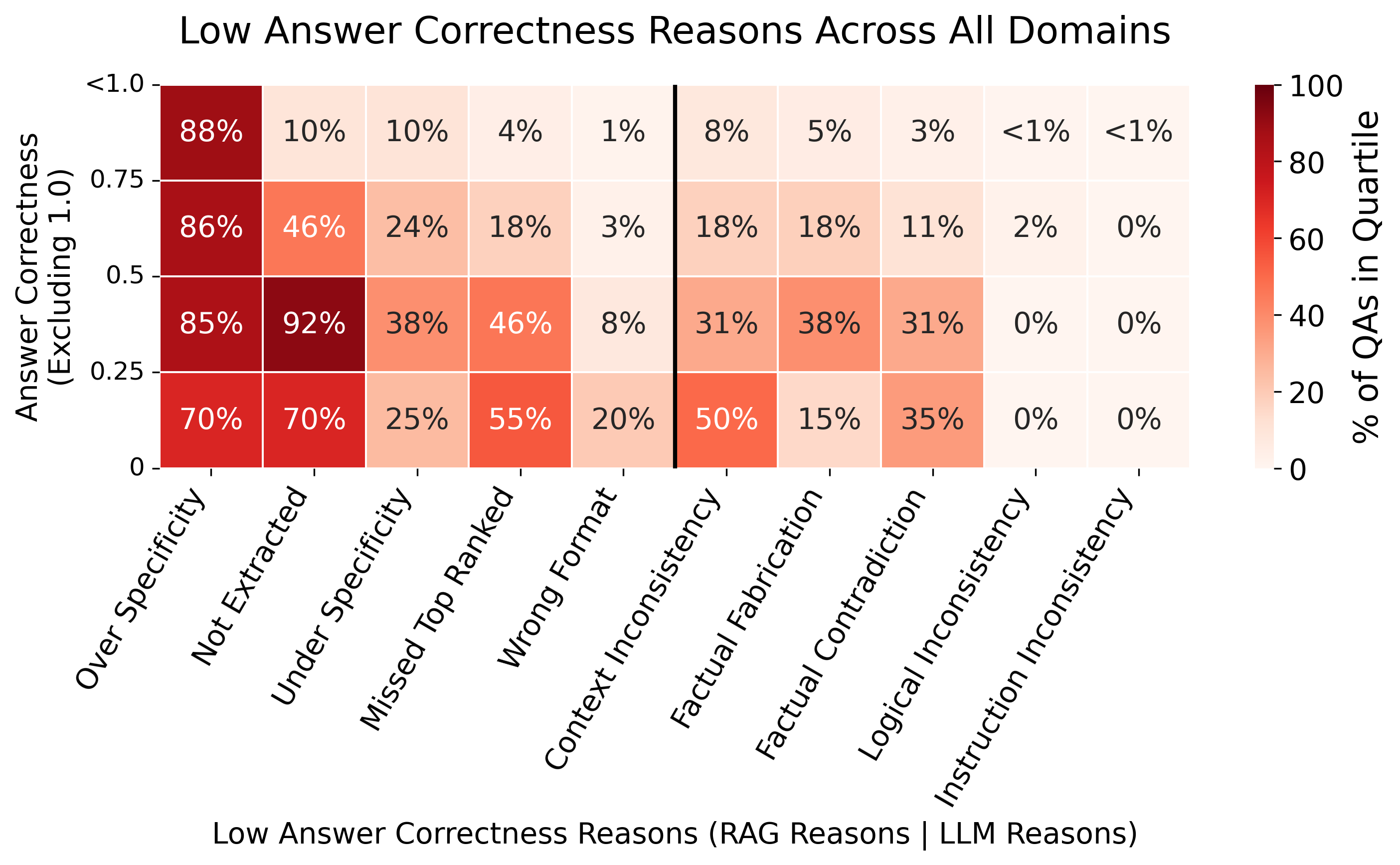}
    \caption{Using GPT-5, we analyze the underlying reasons why Answer Correctness is low by quartile excluding 1.0. We show the number of low Answer Correctness reason as a percentage of the total number of QAs in each quartile. Note that percentages do not sum to 100\% since each QA may exhibit multiple failure reasons. Reasons are grouped by taxonomy with RAG reasons on the left, and LLM reasons on the right.}
    \label{fig:low_correctness_reasons}
\end{figure}

Despite the promising performance of RAG, Answer Correctness remains imperfect, as shown with the non-perfect Answer Correctness in Section \ref{sec:llm_with_rag}. To diagnose the underlying causes, we employ GPT-5 as an LLM-as-a-Judge and analyze failures using a combined taxonomy derived from Barnett et al. \cite{barnett2024sevenfailurepointsengineering} for RAG systems and Huang et al. \cite{Huang_2025} for LLMs. We exclude Missing Content from the RAG taxonomy because our agentic dataset generation pipeline ensures that all QAs are grounded in the source documents. Likewise, we remove Incomplete since the pipeline does not generate multi-part questions. Additionally, we refine Incorrect Specificity into two subcategories—Over Specificity and Under Specificity—to better capture granularity-related errors. Each QA may exhibit multiple failure types.

To assess GPT-5’s reliability as an LLM-as-a-Judge, we manually validate its evaluations on ten QAs from each of the three domain-specific datasets. Across all thirty samples, GPT-5 consistently identifies all correct underlying reasons for low Answer Correctness.

As summarized in Table \ref{tab:failure_counts_grouped}, Over Specificity accounts for the majority of failures. This pattern arises because QA generation relies on a single text chunk as context, leading GPT-4o-mini to produce concise answers constrained by limited information. In contrast, during RAG QA, GPT-4o-mini has access to a larger range retrieved contexts and tends to generate more verbose answers that incorporate all available information. 

For answers scoring above 0.75 in Answer Correctness, as shown in \ref{fig:low_correctness_reasons}, Over Specificity accounts for the vast majority of correctness issues. As Answer Correctness decline, QAs increasingly exhibit multiple contributing factors that lower their overall correctness. More importantly, all other failure reasons still exist, highlighting the weakness in RAG systems.

\subsection{Bias from Dataset Generation LLM}

Recent studies suggest that LLMs often perform better on text they themselves generate \cite{xu2024pride}. To investigate whether this phenomenon introduces bias in RAG performance from our dataset generation pipeline, we construct datasets of 500 questions each using GPT-4o-mini, Gemini-2.5-flash, and Qwen3-30B-A3B-Instruct-2507 within the Army domain. Each dataset is then evaluated with all three models. Retrieval is performed with Qwen3-Embedding-8B, using a chunk size of 800 tokens, 400 token overlap, and top 10 chunk retrieval.

As shown in Table~\ref{fig:llm_with_rag_eval}, we observe minimal evidence of bias from dataset origin. The only instance where Answer Correctness is highest on its own dataset occurs with Qwen3-30B-A3B-Instruct-2507. For Faithfulness, only Gemini-2.5-flash achieves its best score on its own dataset. For Answer Relevancy, none of the models perform best on their respective dataset origins.

\begin{table}[t]
\centering
\caption{LLM Preference for Self-Generated Datasets}
\label{tab:dataset_bias}
\begin{tabular}{@{}lllc@{}}
\toprule
\textbf{Evaluated Model} & \textbf{Metric} & \textbf{Dataset Origin} & \textbf{Score} \\ 
\midrule
\multirow{9}{*}{\textbf{GPT-4o-mini}} & \multirow{3}{*}{Answer Correctness} & GPT-4o-mini & 0.820 \\
 &  & \textbf{Gemini-2.5-flash} & \textbf{0.852} \\
 &  & Qwen3-30B... & 0.827 \\ \cmidrule(l){2-4} 
 & \multirow{3}{*}{Faithfulness} & \textbf{GPT-4o-mini} & \textbf{0.873} \\
 &  & Gemini-2.5-flash & 0.859 \\
 &  & Qwen3-30B... & 0.858 \\ \cmidrule(l){2-4} 
 & \multirow{3}{*}{Answer Relevancy} & GPT-4o-mini & 0.951 \\
 &  & Gemini-2.5-flash & 0.930 \\
 &  & \textbf{Qwen3-30B...} & \textbf{0.960} \\ 
\midrule
\multirow{9}{*}{\textbf{Gemini-2.5-flash}} & \multirow{3}{*}{Answer Correctness} & GPT-4o-mini & 0.852 \\
 &  & Gemini-2.5-flash & 0.874 \\
 &  & \textbf{Qwen3-30B...} & \textbf{0.897} \\ \cmidrule(l){2-4} 
 & \multirow{3}{*}{Faithfulness} & GPT-4o-mini & 0.859 \\
 &  & \textbf{Gemini-2.5-flash} & \textbf{0.942} \\
 &  & Qwen3-30B... & 0.926 \\ \cmidrule(l){2-4} 
 & \multirow{3}{*}{Answer Relevancy} & \textbf{GPT-4o-mini} & \textbf{0.930} \\
 &  & Gemini-2.5-flash & 0.921 \\
 &  & Qwen3-30B... & 0.909 \\ 
\midrule
\multirow{9}{*}{\textbf{Qwen3-30B...}} & \multirow{3}{*}{Answer Correctness} & GPT-4o-mini & 0.876 \\
 &  & Gemini-2.5-flash & 0.859 \\
 &  & \textbf{Qwen3-30B...} & \textbf{0.888} \\ \cmidrule(l){2-4} 
 & \multirow{3}{*}{Faithfulness} & \textbf{GPT-4o-mini} & \textbf{0.888} \\
 &  & Gemini-2.5-flash & 0.879 \\
 &  & Qwen3-30B... & 0.877 \\ \cmidrule(l){2-4} 
 & \multirow{3}{*}{Answer Relevancy} & GPT-4o-mini & 0.929 \\
 &  & \textbf{Gemini-2.5-flash} & \textbf{0.940} \\
 &  & Qwen3-30B... & 0.932 \\ 
\bottomrule
\end{tabular}
\end{table}

\section{Conclusion} \label{conclusion}
We present RAGalyst, an automated human-aligned agentic framework for domain-specific RAG evaluation. We achieve strong human alignment through prompt optimization for Answer Correctness and Answerability. Leveraging these metrics, our framework generates high-quality synthetic QA datasets that outperform both handcrafted benchmarks and RAGAS across all metrics, enabling reliable evaluation of retrieval and generation without human supervision.

Our experiments reveal that RAG performance is highly configuration-dependent, with no universally optimal setup. Embedding model performance varies substantially across domains, often contradicting MTEB rankings. LLM generation performance differs across model families, with closed-source models showing no consistent advantage over open-source alternatives and larger models not consistently outperforming smaller ones. Our low Answer Correctness analysis reveals that RAG systems exhibit imperfect performance, with Incorrect Specificity, Incomplete Extraction, and Context Inconsistency emerging as the three most pressing failure modes.

These findings demonstrate the necessity of a systematic evaluation framework like RAGalyst, which enables practitioners to uncover domain-specific trade-offs and make informed design choices for building reliable RAG systems.



\section*{Acknowledgments}

Research was sponsored by the DEVCOM Analysis Center and was accomplished under Contract Number W911QX-23-D-0009. The views and conclusions contained in this document are those of the authors and should not be interpreted as representing the official policies, either expressed or implied, of the DEVCOM Analysis Center or the U.S. Government. The U.S. Government is authorized to reproduce and distribute reprints for Government purposes notwithstanding any copyright notation herein.

\bibliographystyle{IEEEtran}
\bibliography{references}

\clearpage 

\clearpage
\onecolumn
\section{Supplementary Material}

\subsection{Experimental Figures}

We provide the exact figures from our experimental section \ref{experiments}. We abbreviate the Military Operations domain to Army, Cybersecurity domain to Cyber, and the the Bridge Engineering domain to Eng. 

\subsubsection{Embedding Retrieval Evaluation Across Domains}

The performance results for each embedding model and domain from Figure \ref{fig:embedder_retrieval_eval} are reported in Table \ref{tab:embedder_retrieval_eval}.

\begin{table*}[ht]
    \centering
    \caption{Performance comparison of text embedding models across domains.}
    \label{tab:embedder_retrieval_eval}
    \begin{tabular}{lcccccc}
        \hline
        \textbf{Model} & \multicolumn{3}{c}{\textbf{Recall@10}} & \multicolumn{3}{c}{\textbf{MRR@10}} \\
        \cmidrule(lr){2-4} \cmidrule(lr){5-7}
         & Army & Cyber & Eng & Army & Cyber & Eng \\
        \hline
        text-embedding-3-small & 0.864 & 0.866 & 0.872 & 0.601 & 0.662 & 0.641 \\
        text-embedding-3-large & 0.860 & 0.874 & 0.878 & 0.606 & 0.660 & 0.653 \\
        gemini-embedding-001 & 0.846 & 0.852 & 0.880 & 0.582 & 0.648 & 0.668 \\
        bge-m3 & 0.912 & 0.904 & 0.924 & 0.694 & 0.726 & 0.754 \\
        Qwen3-Embedding-0.6B & 0.904 & 0.892 & 0.926 & 0.701 & 0.723 & 0.745 \\
        Qwen3-Embedding-4B & 0.936 & 0.918 & 0.946 & 0.761 & 0.756 & 0.801 \\
        Qwen3-Embedding-8B & 0.930 & 0.928 & 0.950 & 0.759 & 0.783 & 0.800 \\
        nomic-embed-text-v1 & 0.888 & 0.870 & 0.900 & 0.673 & 0.649 & 0.739 \\
        \hline
    \end{tabular}
\end{table*}

\subsubsection{Domain Specific LLM Generation Evaluation}

The performance results for each LLM and domain from Figure \ref{fig:llm_with_rag_eval} are reported in Table \ref{tab:llm_with_rag_eval}.

\begin{table*}[ht]
    \centering
    \caption{Performance of LLMs across domains for Answer Correctness, Faithfulness, and Answer Relevancy.}
    \label{tab:llm_with_rag_eval}
    \begin{tabular}{lccccccccc}
        \hline
        \textbf{Model} & \multicolumn{3}{c}{\textbf{Answer Correctness}} & \multicolumn{3}{c}{\textbf{Faithfulness}} & \multicolumn{3}{c}{\textbf{Answer Relevancy}} \\
        \cmidrule(lr){2-4} \cmidrule(lr){5-7} \cmidrule(lr){8-10}
         & Army & Cyber & Eng & Army & Cyber & Eng & Army & Cyber & Eng \\
        \hline
        gemma-3-27b-it & 0.86 & 0.84 & 0.86 & 0.93 & 0.94 & 0.93 & 0.91 & 0.91 & 0.92 \\
        Qwen3-30B-A3B-Instruct-2507 & 0.84 & 0.82 & 0.84 & 0.88 & 0.87 & 0.90 & 0.93 & 0.94 & 0.93 \\
        gemini-2.5-flash-lite & 0.85 & 0.84 & 0.86 & 0.93 & 0.92 & 0.94 & 0.94 & 0.96 & 0.94 \\
        gemini-2.5-flash & 0.86 & 0.85 & 0.88 & 0.95 & 0.94 & 0.95 & 0.95 & 0.96 & 0.95 \\
        gemini-2.5-pro & 0.87 & 0.84 & 0.88 & 0.94 & 0.93 & 0.95 & 0.95 & 0.95 & 0.96 \\
        gpt-4o-mini & 0.80 & 0.79 & 0.82 & 0.88 & 0.90 & 0.91 & 0.94 & 0.95 & 0.94 \\
        gpt-4.1-nano & 0.83 & 0.81 & 0.85 & 0.89 & 0.90 & 0.92 & 0.93 & 0.93 & 0.95 \\
        gpt-4.1-mini & 0.85 & 0.84 & 0.86 & 0.91 & 0.94 & 0.93 & 0.92 & 0.94 & 0.94 \\
        gpt-4.1 & 0.84 & 0.82 & 0.86 & 0.92 & 0.91 & 0.93 & 0.94 & 0.95 & 0.95 \\
        \hline
    \end{tabular}
\end{table*}

\subsubsection{Domain Specific Retrieved Chunks Evaluation}

The performance results of Gemma3-4B on Answer Correctness, Faithfulness, and Answer Relevancy on all domains from Figure \ref{fig:chunks_ablation} are reported in Table \ref{tab:chunks_ablation}

\begin{table*}[ht]
    \centering
    \caption{Performance of domains across varying numbers of chunks.}
    \label{tab:chunks_ablation}
    \begin{tabular}{llcccccccccc}
        \hline
        \textbf{Metric} & \textbf{Domain} & \textbf{1} & \textbf{2} & \textbf{3} & \textbf{4} & \textbf{5} & \textbf{6} & \textbf{7} & \textbf{8} & \textbf{9} & \textbf{10} \\
        \hline
        \multirow{3}{*}{Faithfulness} 
         & Army & 0.880 & 0.900 & 0.874 & 0.884 & 0.863 & 0.876 & 0.864 & 0.875 & 0.865 & 0.878 \\
         & Cyber & 0.857 & 0.893 & 0.889 & 0.893 & 0.867 & 0.883 & 0.879 & 0.876 & 0.886 & 0.875 \\
         & Eng & 0.905 & 0.901 & 0.896 & 0.894 & 0.887 & 0.891 & 0.887 & 0.891 & 0.863 & 0.896 \\
        \hline
        \multirow{3}{*}{Answer Relevancy} 
         & Army & 0.813 & 0.835 & 0.853 & 0.841 & 0.871 & 0.856 & 0.852 & 0.854 & 0.854 & 0.864 \\
         & Cyber & 0.860 & 0.868 & 0.871 & 0.877 & 0.884 & 0.899 & 0.893 & 0.892 & 0.901 & 0.877 \\
         & Eng & 0.846 & 0.831 & 0.847 & 0.866 & 0.868 & 0.873 & 0.859 & 0.878 & 0.872 & 0.873 \\
        \hline
        \multirow{3}{*}{Answer Correctness} 
         & Army & 0.792 & 0.808 & 0.810 & 0.820 & 0.820 & 0.812 & 0.794 & 0.808 & 0.791 & 0.792 \\
         & Cyber & 0.776 & 0.800 & 0.797 & 0.792 & 0.784 & 0.782 & 0.789 & 0.786 & 0.779 & 0.769 \\
         & Eng & 0.787 & 0.812 & 0.808 & 0.812 & 0.817 & 0.793 & 0.792 & 0.801 & 0.790 & 0.788 \\
        \hline
    \end{tabular}
\end{table*}

\newpage

\subsection{LLM Prompts}

\subsubsection{Hand-Crafted Answer Correctness Prompt}

The following is our hand-crafted prompt for Answer Correctness.

\begin{tcolorbox}[colback=gray!5!white, colframe=gray!60!black, title={Evaluation Prompt}]
    \small
    You will be given a student answer and a ground truth.\\
    
    Your task is to evaluate the student answer by comparing it with the ground truth.
    Give your evaluation on a scale of 0.0 to 1.0, where 0.0 means that the answer is completely unrelated to the ground truth, and 1.0 means that the answer is completely accurate and aligns perfectly with the ground truth. \\
    
    For instance,\\
    correctness\_score: 0.0 -- The answer is completely unrelated to the ground truth.\\
    correctness\_score: 0.3 -- The answer has minor relevance but does not align with the ground truth.\\
    correctness\_score: 0.5 -- The answer has moderate relevance but contains inaccuracies.\\
    correctness\_score: 0.7 -- The answer aligns with the reference but has minor errors or omissions.\\
    correctness\_score: 1.0 -- The answer is completely accurate and aligns perfectly with the ground truth.\\
    
    You must provide values for correctness\_score: in your answer.\\
    
    Now here is the student answer and the ground truth.
\end{tcolorbox}

\newpage

\subsubsection{Prompt-Optimized Answer Correctness Prompt}

The following is the MIPROv2 and LabeledFewShot optimized Answer Correctness prompt.

\begin{tcolorbox}[colback=gray!5!white, colframe=gray!60!black, title={Evaluation Prompt}]
    \small
    \texttt{
    \{\\
    \ \ \ \ "response": "8 rockets fired from Gaza into southern Israel; none hurt",\\
    \ \ \ \ "reference": "Ten rockets from Gaza land in southern Israel; none hurt",\\
    \ \ \ \ "correctness\_score": 0.7\\
    \ \ \},\\[0.5em]
    \ \ \{\\
    \ \ \ \ "response": "A person plays a keyboard.",\\
    \ \ \ \ "reference": "Someone is playing a keyboard.",\\
    \ \ \ \ "correctness\_score": 1.0\\
    \ \ \},\\[0.5em]
    \ \ \{\\
    \ \ \ \ "response": "What isn't how what was sold?",\\
    \ \ \ \ "reference": "It's not how it was sold, gb.",\\
    \ \ \ \ "correctness\_score": 0.3\\
    \ \ \},\\[0.5em]
    \ \ \{\\
    \ \ \ \ "response": "Jaya Prada all set to join BJP",\\
    \ \ \ \ "reference": "Jaya Prada likely to join BJP, Amar Singh to decide for her",\\
    \ \ \ \ "correctness\_score": 0.8\\
    \ \ \},\\[0.5em]
    \ \ \{\\
    \ \ \ \ "response": "Israel strikes Syria as tensions rise on weapons",\\
    \ \ \ \ "reference": "Air strikes wound civilians in Syria's Deraa",\\
    \ \ \ \ "correctness\_score": 0.4\\
    \ \ \},\\[0.5em]
    \ \ \{\\
    \ \ \ \ "response": "The issue has been resolved, Marlins President David Samson said through a club spokesman.",\\
    \ \ \ \ "reference": "The Marlins only said: \"The issue has been resolved.\"",\\
    \ \ \ \ "correctness\_score": 0.6\\
    \ \ \},\\[0.5em]
    \ \ \{\\
    \ \ \ \ "response": "Typhoon survivors raid Philippine stores",\\
    \ \ \ \ "reference": "Typhoon Bopha kills 15 in S. Philippines",\\
    \ \ \ \ "correctness\_score": 0.2\\
    \ \ \},\\[0.5em]
    \ \ \{\\
    \ \ \ \ "response": "three little boys cover themselves with bubbles.",\\
    \ \ \ \ "reference": "Three children standing by a pool are covered in foam bubbles.",\\
    \ \ \ \ "correctness\_score": 0.8\\
    \ \ \}
    }

    You are a language assessment evaluator. You will be given a student answer and a ground truth response. Your task is to evaluate the student answer by comparing it with the ground truth and provide a similarity score on a scale of 0.0 to 1.0. A score of 0.0 indicates that the answer is completely unrelated to the ground truth, while a score of 1.0 indicates that the answer is completely accurate and aligns perfectly with the ground truth. Please include the evaluation in the format: correctness\_score: [score]. \\ \\
    
    Now here is the student answer and the ground truth.
\end{tcolorbox}

\newpage
\subsubsection{Answerability}

The following is our hand-crafted prompt for Answerability.

\begin{tcolorbox}[colback=gray!5!white, colframe=gray!60!black, title={Evaluation Prompt}]
    \small
    You will be given a context and a question.\\

    Your task is to determine if the question is clearly and unambiguously answerable using only the given context.
    - If the context contains **all** the necessary information to answer the question **without making assumptions** or using **any external knowledge**, then the groundedness is 1.
    - Otherwise, if any key information is **missing**, ambiguous, or requires inference beyond what is stated, then the groundedness is 0.\\

    You MUST provide values for 'answerability\_flag:' in your answer.\\ 

    Use only the provided context. Do not use prior knowledge, common sense, or information not explicitly contained in the context.
\end{tcolorbox}

\end{document}